\documentclass[11pt]{article}

\usepackage[preprint]{acl}

\usepackage{times}
\usepackage{latexsym}
\usepackage[whole]{bxcjkjatype}

\usepackage[T1]{fontenc}

\usepackage[utf8]{inputenc}

\usepackage{microtype}

\usepackage{inconsolata}

\usepackage{graphicx}
\usepackage{subcaption}
\usepackage{enumitem}
\usepackage{pifont}
\usepackage[most]{tcolorbox}
\usepackage{longtable}
\usepackage{array}
\usepackage{xltabular}
\usepackage{refcount}

\usepackage{amsmath}
\usepackage{amsthm}
\usepackage{booktabs}
\usepackage{multirow}
\usepackage{caption}
\usepackage{amssymb}
\usepackage[table,xcdraw,dvipsnames]{xcolor}

\usepackage{fontawesome5}
\usepackage{minted}

\lstdefinelanguage{json}{
    basicstyle=\ttfamily\small,
    numbers=left,
    numberstyle=\tiny,
    stepnumber=1,
    breaklines=true,
    showstringspaces=false,
    string=[s]{"}{"},
}

\title{\textsc{HalluCiteChecker}: A Lightweight Toolkit for Hallucinated Citation Detection and Verification in the Era of AI Scientists}

\author{Yusuke Sakai, \;\;
        Hidetaka Kamigaito, \;\;
        Taro Watanabe \\
  Nara Institute of Science and Technology (NAIST), Japan  \\
  \texttt{\{sakai.yusuke.sr9, kamigaito.h, taro\}@is.naist.jp}}

\begin{document}
\maketitle
\begin{abstract}

We introduce \textsc{HalluCiteChecker}, a toolkit for detecting and verifying hallucinated citations in scientific papers.
While AI assistant technologies have transformed the academic writing process, including citation recommendation, they have also led to the emergence of hallucinated citations that do not correspond to any existing work.
Such citations not only undermine the credibility of scientific papers but also impose an additional burden on reviewers and authors, who must manually verify their validity during the review process.
In this study, we formalize hallucinated citation detection as an NLP task and provide a corresponding toolkit as a practical foundation for addressing this problem. Our package is lightweight and can perform verification in seconds on a standard laptop.
It can also be executed entirely offline and runs efficiently using only CPUs.
We hope that \textsc{HalluCiteChecker} will help reduce reviewer workload and support organizers by enabling systematic pre-review and publication checks.
Our code is released under the Apache 2.0 license on GitHub\footnote{\faGithub \ \ : \href{https://github.com/yusuke1997/HalluCiteChecker}{\texttt{yusuke1997/HalluCiteChecker}}} and is distributed as an installable package via PyPI\footnote{\faPython \ \ : \href{https://pypi.org/project/hallucitechecker/}{\texttt{pip install hallucitechecker}}}. A demonstration video is available on YouTube\footnote{\label{footnote:youtube}\faYoutube \ : \url{https://youtu.be/WG3WXgH2pok}}.
\end{abstract}

\section{Introduction}

\begin{figure}[t]
    \centering
    \includegraphics[width=0.865\linewidth]{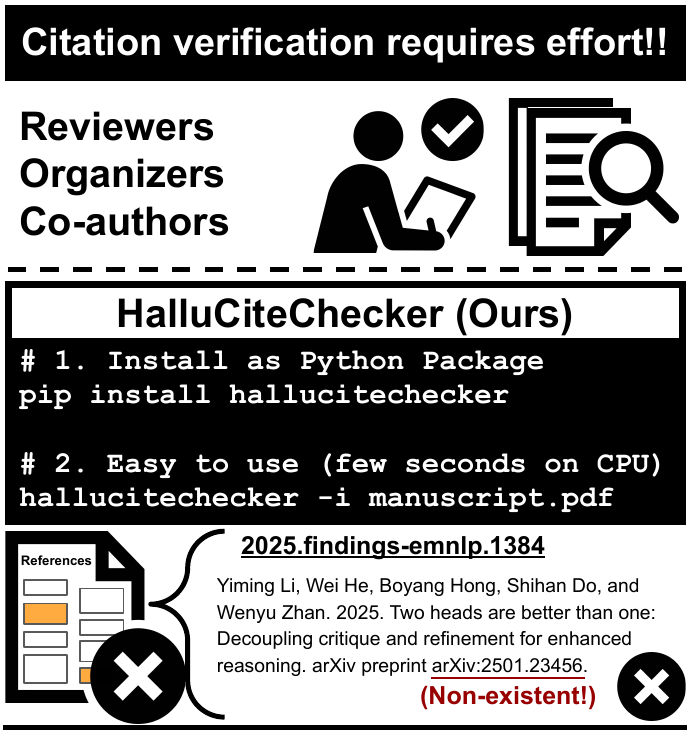}
    \caption{Illustration of the usage scenario of \textsc{HalluCiteChecker}. Hallucinated citations are often disguised among many correct citations, making manual verification labor-intensive. 
    \textsc{HalluCiteChecker} is easy to install and works out-of-the-box on a laptop. It highlights candidate hallucinated citations, enabling efficient verification, promoting author awareness, and reducing reviewer workload. In the example paper~\cite{ji-lu-2025-reflair}, more than ten citations were identified.}
    \label{fig:overview}
\end{figure}

\begin{figure*}[t]
    \centering
    \includegraphics[width=\linewidth]{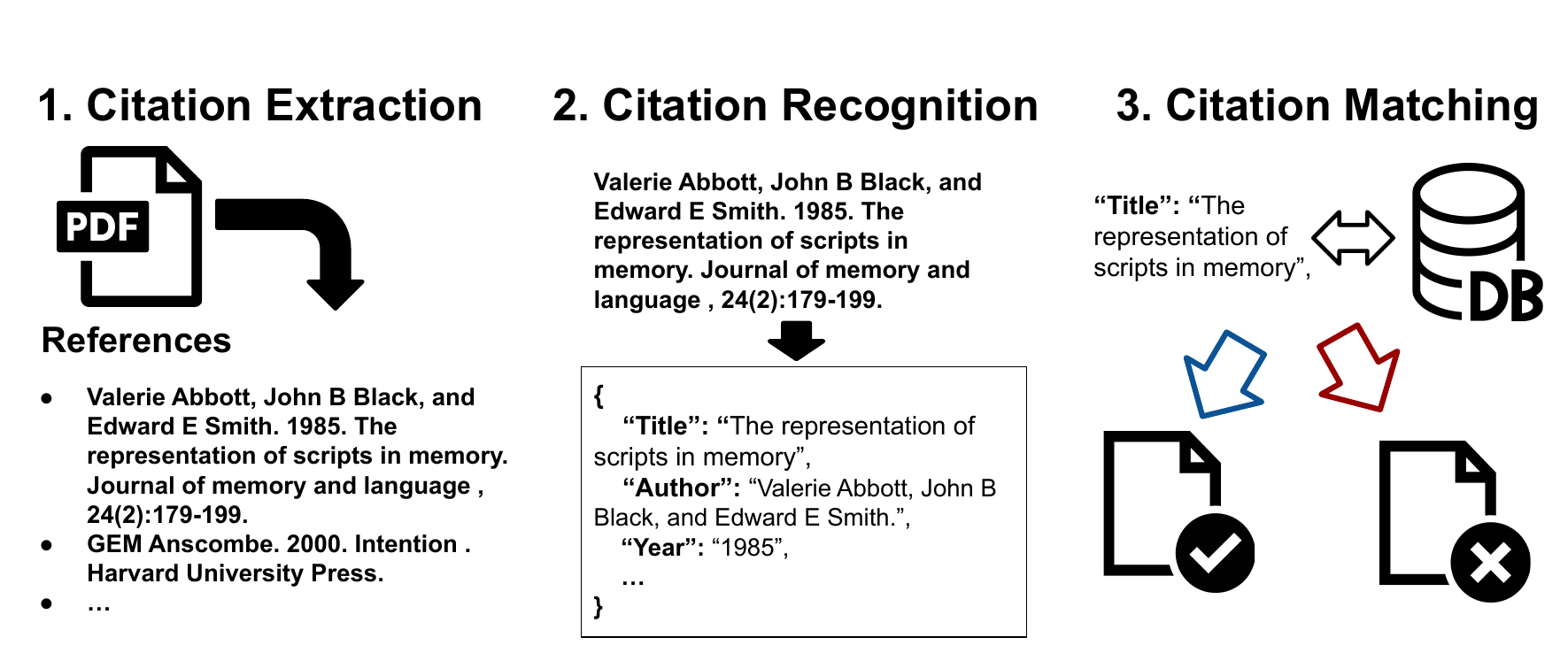}
    \caption{Overview of the hallucinated citation detection task, which consists of three subtasks: Citation Extraction (Section~\ref{sec:task-extraction}), Citation Recognition (Section~\ref{sec:task-recognition}), and Citation Matching (Section~\ref{sec:task-matching}).}
    \label{fig:task-definition}
\end{figure*}

Citations are a fundamental component of scientific papers~\cite{wang2021science}. They serve to substantiate claims with evidence, acknowledge prior work, and clarify the positioning and novelty of new contributions within a research area. Beyond supporting scientific arguments, citations also function as a key quantitative metric for evaluating scientific impact. Citation counts are widely used to assess the influence of papers, researchers, and publication venues, forming the basis of metrics such as the h-index and impact factor. As a result, citations serve not only as scholarly references but also as a mechanism for allocating academic credit.

Recently, reflecting the advancement of AI for Science and AI Scientist technologies, writing assistance powered by large language models (LLMs) has transformed the academic writing process, including automated drafting, paraphrasing, and citation recommendation. These tools improve productivity and accessibility, particularly for non-native English speakers, junior researchers, and students. However, they have also introduced new risks.
Hallucinated citations, which are references that do not correspond to any real work or contain incorrect bibliographic information, have begun to appear increasingly in papers under review, preprints, and even peer-reviewed publications. Recent reports indicate that their prevalence has increased rapidly~\cite{sakai2026hallucitationmattersrevealingimpact}. This trend poses a serious threat to the integrity and reliability of scientific communication, which has traditionally relied on an implicit assumption of good faith, where authors accurately cite existing and relevant work\footnote{Erroneous citations have long been known to exist in scientific literature~\cite{wang2021science, Errors-in-Bibliographic-Citations, xu2026ghostcitelargescaleanalysiscitation}. While such errors were traditionally attributed to human mistakes, the LLM era has seen a rapid increase in hallucinated citations, including fabrication of non-existent works~\cite{xu2026ghostcitelargescaleanalysiscitation, Oladokun02012025, Walters2023}. In addition, minor inconsistencies, such as slight variations in bibliographic details, have also become increasingly common~\cite{bienz2026casemysteriouscitations}. Regardless of their origin, it is desirable to systematically detect them.}.

The growing presence of hallucinated citations has shifted additional verification responsibilities onto reviewers, who must now manually confirm the validity of cited references to maintain publication quality.
This verification process imposes a significant and unilateral burden on reviewers and publication organizers.
Consequently, papers containing hallucinated citations may face desk rejection or other corrective actions.
Despite these efforts, systematic investigations have revealed that papers containing hallucinated citations have been accepted to several venues.
These findings highlight the limitations of manual verification and underscore the urgent need for automated detection methods.

To address this challenge, we formalize hallucinated citation detection as a novel NLP task and introduce \textsc{HalluCiteChecker}, a lightweight toolkit for verifying and detecting hallucinated citations in scientific papers, as shown in Figure~\ref{fig:overview}.
Our system is designed to be practical and accessible: it can perform verification within seconds on a standard laptop, runs entirely offline, and requires only a CPU without external communication or specialized hardware.
This enables seamless integration into existing workflows, including pre-submission checks by authors, reviewer assistance during peer review, and automated screening by conference organizers and publishers.
By providing an efficient and deployable solution for hallucinated citation detection, \textsc{HalluCiteChecker} aims to reduce reviewer workload, improve publication reliability, and help preserve the integrity of scientific communication in the era of AI-assisted research.

\section{Task Definition}
\label{sec:task-definition}

We define the Hallucinated Citation Detection task as a structured task consisting of three subtasks: Citation Extraction, Citation Recognition, and Citation Matching. Figure~\ref{fig:task-definition} shows the overview of these subtasks. This decomposition clarifies the challenges involved in hallucinated citation detection and highlights opportunities for systematic improvement and future development.

\subsection{Citation Extraction}
\label{sec:task-extraction}

Scientific papers are typically distributed as structured documents, e.g., PDF files\footnote{Scientific research is also presented in various formats, including physical printouts and HTML pages such as blog posts. In this paper, we focus on PDF documents for simplicity of exposition. These alternative formats are merely variations in input format. Moreover, reliably extracting complete and accurate citation information remains challenging, despite significant progress in document parsing research.}. Under this constraint, the first step is to extract citation information from the document. Citation extraction involves identifying and parsing bibliographic entries from raw documents, which presents challenges due to variations in layout, formatting, and document quality.
This subtask offers several directions for improvement and extension. Recognition accuracy and efficiency can be improved through parsing, OCR, and post-correction. Moreover, it is necessary to isolate citation-specific information, such as identifying reference sections and separating individual bibliographic records.

In addition, this subtask can be extended to extract peripheral information, including structural information, e.g., bounding-box coordinates, and contextual information associated with citations. Such extensions enable a structural understanding and visualization of citation usage, and provide additional signals for downstream recognition and verification, including assessing whether citations are appropriate and valid. In summary, this subtask focuses on extracting citation-related information, including bibliographic entries and associated structural and contextual metadata, from documents.

\subsection{Citation Recognition}
\label{sec:task-recognition}

Citation information is typically represented as inline text within documents. Therefore, it is necessary to identify and classify individual citation components, such as the title, author names, publication year, venue, and page numbers. Citation styles vary widely, including standard formats such as APA and MLA, as well as conference- and journal-specific formatting conventions. In addition, citation entries may contain missing or incomplete fields. Consequently, naive rule-based extraction approaches are insufficient in real-world scenarios.

This subtask aims to identify the semantic structure of citation text in the wild by determining which spans correspond to specific bibliographic components. This can be regarded as a specialized form of named entity recognition (NER) task, where the goal is to extract structured bibliographic fields from raw citation strings. Although existing NER techniques can be adapted for scientific documents, challenges remain due to the large number of entity types and the diversity of citation formats encountered across venues and disciplines.

\subsection{Citation Matching}
\label{sec:task-matching}

Given structured citation information, we determine whether each citation is valid by matching it against bibliographic databases, such as local databases and web-based scholarly search engines. While it is possible to use various citation fields for this purpose, prior studies on hallucinated citation verification have shown that title information serves as the reliable matching key~\cite{sakai2026hallucitationmattersrevealingimpact, bienz2026casemysteriouscitations, xu2026ghostcitelargescaleanalysiscitation}\footnote{One possible reason is that author names often appear in abbreviated or truncated forms, such as the use of initials, omission of middle names, or the use of ``et al.'' to represent multiple authors. Additionally, venue names are frequently abbreviated, and bibliographic fields such as page numbers may be incomplete or inconsistently formatted. While it is possible to require identifiers such as DOIs or URLs, these identifiers are not consistently available for all citations. Moreover, DOI resolution pages vary in structure and formatting, and retrieving reliable metadata from them introduces additional challenges. In contrast, title information is typically more distinctive, stable, and presented in its full form, making it a practical and reliable key for citation verification.}.
Therefore, in this paper, we primarily focus on title-based citation matching. Note that if structured bibliographic data, such as BibTeX entries, are available, citation verification can be conducted directly starting from this subtask. However, such structured metadata is rarely available in practice. Therefore, it is desirable to consider the entire pipeline, including citation extraction, recognition, and matching.

\section{\textsc{HalluCiteChecker}: Principles}
\label{sec:design-principles}

We design our toolkit, \textsc{HalluCiteChecker}, with the following five design principles in mind:

\begin{description}[leftmargin=0.5em, itemsep=0em, topsep=0.5em]

\item[Easy to Use:]
The toolkit should be easy to install and use, avoiding complex setup procedures. To achieve this, we provide the toolkit as a Python package that can be installed with a single command and executed through a one-line command. Our goal is to enable out-of-the-box functionality without requiring additional configuration, similar to widely used tools such as \textsc{ACL pubcheck}\footnote{\label{footnote:aclpubcheck}\url{https://github.com/acl-org/aclpubcheck}}.

\item[Lightweight:]
The toolkit should be practical and efficient in typical usage environments. Therefore, it is designed to run efficiently even on standard laptops using only CPU resources and within practical time constraints.

\item[Offline:]
For instance, during the peer-review process, uploading manuscripts to external APIs or web services may introduce risks of data leakage and violate confidentiality policies. To address these concerns, the toolkit is designed to function fully offline without relying on external services. While optional external integrations may be supported, the core functionality remains fully operational in isolated, in-house environments.

\item[Non-Generative AI:]
Generative AI tools, such as LLMs, may be restricted or prohibited under certain peer-review policies\footnote{E.g., the ICML 2026: \url{https://icml.cc/Conferences/2026/LLM-Policy}, and the ACL Rolling Review Policy: \url{https://aclrollingreview.org/reviewerguidelines\#q-can-i-use-generative-ai} [Accessed on Apr. 27th, `26]}. Reliance on such tools could introduce compliance, reproducibility, and hallucination concerns. Hence, the core implementation avoids the use of generative AI and instead relies on standard NLP and ML techniques.

\item[Simple Implementation:]
It follows a minimalist design to facilitate maintainability and extensibility. By providing well-decomposed components and a clear implementation structure with explicit separation between base classes and specific implementations, and enabling easy following of the data and processing flow, the toolkit supports easy integration of new methods with minimal effort.

\end{description}

\begin{lstlisting}[float=!t, caption = Pseudo-code of \textsc{HalluCiteChecker} to illustrate our design. It is simplified for this paper based on the base-class components. Each class corresponds to a subtask in our pipeline. Therefore{,} the system works as long as the derived codes comply with the input and output interfaces., label = lst:design]
from abc import ABC, abstractmethod
from typing import List
from dataclasses import dataclass, field

'### Section 4.4: Dataclass `Citation` ###'
@dataclass(slots=True)
class Citation:
  raw_text: str
  bboxes: list = field(default_factory=list)
  title: str = ""
  author: str = ""
  ...
  
'### Section 4.1: Citation Extractor ###'
class BaseCitationExtractor(ABC):
  @abstractmethod
  def extract(path: str) -> List[Citation]:
    # Extract each citation from the PDF and return it as a List[Citation].
    pass
    # [Citation(raw_text, bboxes, ...), ...]
  ...
  
'### Section 4.2: Citation Recognizer ###'
class BaseCitationRecognizer(ABC):
  @abstractmethod
  def parse(citation: Citation) -> Citation:
    # Parse the citation string and store the bibliographic information.
    pass
    # Citation(raw_text='', bboxes= [...], title='', author='', ...)
    
  def batch_parse(citations: List[Citation], ...) -> List[Citation]:
    # In practice, more batchfied them.
    return [self.parse(c) for c in citations]

'### Section 4.3: Citation Matcher ###'
class BaseCitationMatcher(ABC):
  @abstractmethod
  def is_match(self, c: Citation) -> bool:
    # Matching via databases or APIs.
    # If necessary, store the DBs in __init__
    pass

  def verify(citations: List[Citation]) -> List[Citation]:
    # If you want to batch process or return rich information, override `verify`.
    return [c for c in citations if not self.is_match(c)]

\end{lstlisting}

\section{\textsc{HalluCiteChecker}: System Details}

We construct a framework where the three subtasks defined in Section~\ref{sec:task-definition} are implemented as individual components, as shown in Listing~\ref{lst:design}. Each class corresponds to a key subtask. This modular design supports easy extension and enables flexible integration of new methods for future research.
The pipeline is controlled through a unified \texttt{Citation} dataclass, which manages data flow across components and provides a clear and structured workflow. Based on this design, we primarily describe the default out-of-the-box implementations.

\subsection{Citation Extractor}
\label{sec:citation-extractor}

Given a PDF document as input, this component extracts the raw text of each citation and returns a list of \texttt{Citation} objects. We use \texttt{Docling}~\cite{auer2024doclingtechnicalreport} to perform PDF parsing using only CPU resources\footnote{\texttt{Docling} provides optional OCR and LLM-based correction features that may utilize GPU resources. However, we use only its internal CPU-based parsing module, \textit{docling-parse}, as a front-end tool for PDF parsing.}. 
Next, we heuristically identify and extract the reference section, and then apply post-processing to reconstruct citation strings that may be split across page breaks or line boundaries. This process yields citation information as one-line raw text strings. In addition, structural metadata such as bounding box coordinates is obtained. The \texttt{Docling} is available under the MIT License.

\subsection{Citation Recognizer}
\label{sec:citation-recognizer}

Given citation strings obtained from Section~\ref{sec:citation-extractor}, this component extracts the title information. However, heuristic approaches based on regular expressions have inherent limitations due to variations in citation formats~\cite{bienz2026casemysteriouscitations}. Therefore, citation recognition is formulated as a sequence labeling task similar to NER, where each token in the citation string is tagged to identify structured bibliographic fields, including the title.

We employ DeLFT~\cite{DeLFT} to extract structured bibliographic fields, including the title, from citation strings\footnote{Prior analysis work~\cite{sakai2026hallucitationmattersrevealingimpact, xu2026ghostcitelargescaleanalysiscitation} has used GROBID~\cite{GROBID}, a de facto tool for scholarly document parsing. However, GROBID is implemented in Java and typically requires setting up a Docker environment, which does not align with our Easy-to-Use design principle, described in Section~\ref{sec:design-principles}. Therefore, to meet our requirements, we use DeLFT, a Python-based reimplementation of GROBID's sequence-labeling models.}.
Specifically, we use a pretrained \texttt{BidLSTM\_CRF\_FEATURES} model~\cite{lample-etal-2016-neural, chiu-nichols-2016-named} with \texttt{glove840B} embeddings~\cite{pennington-etal-2014-glove}. Given a citation string, the model performs sequential labeling with 18 entity tags\footnote{18 entities are as follows: author, booktitle, collaboration, date, editor, institution, issue, journal, location, note, pages, publisher, pubnum, series, tech, title, volume, and web.}, from which the title field is extracted. The tools used in this component are available under the Apache 2.0 License.

\subsection{Citation Matcher}
\label{sec:citation-matcher}

Given the extracted title information, this component verifies whether the cited paper exists by matching it against bibliographic databases.
Based on~\citet{sakai2026hallucitationmattersrevealingimpact, xu2026ghostcitelargescaleanalysiscitation}, we employ character-level fuzzy matching on citation titles using similarity scores based on the normalized Levenshtein distance implemented in the \texttt{RapidFuzz} library~\cite{max_bachmann_2025_15133267}, with a threshold of 0.9. 
We also include bibliographic databases derived from ACL Anthology, arXiv, and DBLP as part of the default setting, following \citet{sakai2026hallucitationmattersrevealingimpact}.

Although Listing~\ref{lst:design} shows a simplified implementation that returns a boolean verification result, the system can be easily extended to return similarity scores or additional metadata by overriding the \texttt{verify()} function.
While external APIs can be useful for this task, to comply with our design principles stated in Section~\ref{sec:design-principles}, we do not include external API access in the core module. Instead, external API access is treated as optional, and citation matching is primarily performed against internal bibliographic databases. The libraries used in this component are available under the MIT License.

\subsection{Unified Dataclass: \texttt{Citation}}

As shown in Listing~\ref{lst:design}, each component works by adding information to a unified \texttt{Citation} dataclass, which serves as the central data structure for managing all citation-related information. In particular, representing each citation as a single \texttt{Citation} object from the beginning eliminates the need to assign or manage sequential identifiers, resulting in a simpler and more maintainable implementation. This design also enables peripheral information obtained in earlier stages, such as bounding box coordinates, to be preserved and accessed by downstream components without requiring intermediate steps to manage such metadata.

Furthermore, processing decisions are made based on the presence of required fields, and undefined attributes are treated as errors. This design enforces a predefined and type-safe data structure, enabling robust handling of incomplete or exceptional cases without disrupting the overall workflow. Moreover, the dataclass provides utility methods such as \texttt{from\_dict} and \texttt{display}, which encapsulate data formatting and representation within the dataclass itself. This explicit separation between data representation and pipeline processing facilitates tracing and debugging across pipeline stages while minimizing implementation complexity.

\subsection{Logistics and Infrastructures}

\paragraph{Pretrained weights and databases.}

While the \textsc{HalluCiteChecker} codebase can be easily installed via PyPI, certain resources used in Sections~\ref{sec:citation-recognizer} and~\ref{sec:citation-matcher} are too large to be distributed directly within the package. Although users can initialize and update these resources locally, we provide out-of-the-box access to pretrained weights and bibliographic databases to ensure ease of use and efficient loading. To minimize loading latency, the \texttt{glove840B} embeddings are distributed through the Hugging Face Hub~\cite{wolf-etal-2020-transformers} in Lightning Memory-Mapped Database (LMDB) format, enabling efficient memory-mapped access. The citation databases are distributed using Hugging Face Datasets~\cite{lhoest-etal-2021-datasets}, which leverage Apache Arrow for fast loading and efficient storage. These databases are automatically versioned and updated without requiring manual intervention.

\paragraph{Developer tools.}

We provide developer tools to facilitate performance measurement and system evaluation. This includes a stopwatch utility for measuring processing time~\cite{deguchi-etal-2024-mbrs} %
and profiling tools that export execution environment information such as memory usage.

\paragraph{PDF highlighting for review support.}

We support the processing of multiple PDF documents and provide functionality to generate highlighted output PDFs indicating detected hallucinated citation candidates. 
To support fair and consistent review and publication checks, the toolkit allows fixed database versions to be used and shared among reviewers, organizers, and authors. This ensures reproducible verification results and facilitates efficient consensus-building for handling hallucinated citations, including desk rejection decisions. If false positives occur, authors can provide supporting documentation to justify the citation, thereby preventing inappropriate desk rejection. This enables more efficient and transparent verification while maintaining fairness in the review process.

\begin{lstlisting}[language=sh, float=!t, caption = CLI-based usage of \textsc{HalluCiteChecker}., label = lst:cli]
hallucitechecker -i ok.pdf hallucinated.pdf \
                [-o outputs] ...

# All Clear!  # -> ok.pdf
# {'author': 'Yiming Li, Wei He, ...,
#   'title': 'Two heads are better than...'
# }, ...      # -> hallucinated.pdf

\end{lstlisting}

\begin{lstlisting}[float=!t, caption = An example of the current implementation demonstrating the use of custom databases or APIs., label = lst:api]
from HalluCiteChecker import *

doc_extractor = DoclingCitationExtractor()
doc_extractor.initialize()
citation_parser = DelftCitationRecognizer()

# Use your own database: (csv, tsv, Datasets)
# Require fields: title [, id]
db1 = FuzzCitationMatcher(db_path="...", score_cutoff=70, ...)
db2 = FuzzCitationMatcher(db_path="...", ...)
api_db = YourCustomAPICitationMatcher(...)

path ='your_manuscripts.pdf'
# Section 4.1: Citation Extractor
refs = doc_extractor.extract_references(path)
# Section 4.2: Citation Recognizer
citations = citation_parser.parse_batch(refs)
# len(citations) -> 50

# Section 4.3: Citation Matcher (customized)
citations = db1.verify(citations)
citations = db2.verify(citations)
citations = api_db.verify(citations)
# len(citations) -> 1 (hallucinated citation)
\end{lstlisting}

\section{Try!: \texttt{pip install hallucitechecker}}

\paragraph{Case 1: CLI-based verification.}

Listing~\ref{lst:cli} shows example usage via the command-line interface (CLI), which enables verification of multiple PDF files as input. The toolkit displays \textit{All Clear!} if no hallucinated citations are detected. Otherwise, detected candidates are reported in the CLI output.
The toolkit also provides optional features. For example, by setting an output directory, highlighted PDFs can be generated with detected citation candidates visually marked, as illustrated in Figure~\ref{fig:overview}.

\paragraph{Case 2: Python API-based customization.}

As shown in Listing~\ref{lst:api}, each component can be used independently via the Python API, enabling flexible customization. The toolkit follows a one-function, one-object design principle, allowing components to be easily composed in multi-stage pipelines. For example, users can apply multiple databases sequentially by chaining instances of the component.

\section{Evaluation}
\label{sec:evaluation}

\begin{table}[t]
    \centering
    \small
    \setlength{\tabcolsep}{1.75pt}
    \begin{tabular}{@{}lrrrr@{}} 
    \toprule
       Time  & Extractor & Recognizer & Matcher & Total \\ \midrule
    \textit{MacBook Pro} \\
     msec/paper     & 6206.8ms & 215.9ms & 798.1ms & 7220.9ms \\ 
     msec/citation  &  121.7ms &   4.2ms &  15.6ms & 141.6ms \\
     Total (sec)    &  620.7s &  21.6s &  79.8s & 722.1s \\
     \midrule
     \textit{MacBook Air} \\
     msec/paper     & 14611.9ms & 386.2ms & 1363.4ms & 16361.7ms \\ 
     msec/citation  &  286.5ms  &   7.6ms &   26.7ms &   320.8ms \\
     Total (sec)    & 1461.2s &  38.6s &  136.3s &  1636.2s \\
     \midrule
     \textit{WSL} \\
     msec/paper     & 31970.5ms  &  329.4ms & 2366.4ms & 34666.5ms \\ 
     msec/citation  &   626.9ms &    6.5ms &   46.4ms &   679.7ms \\
     Total (sec)    & 3197.1s &   32.9s &  236.6s &  3466.7s \\
     \bottomrule
    \end{tabular}
    \caption{Processing performance comparison of \textsc{HalluCiteChecker}. We report the average processing time per paper and per citation in milliseconds.}
    \label{tab:speed}
\end{table}

\textsc{HalluCiteChecker} is primarily designed for efficiency and usability. Regarding detection performance, we verified that the toolkit successfully detects hallucinated citations in papers from the EMNLP 2025 Main track reported in \citet{sakai2026hallucitationmattersrevealingimpact}. This indicates that it achieves at least comparable recall to that of prior analyses. Our \href{https://youtu.be/WG3WXgH2pok}{YouTube video}\footref{footnote:youtube} demonstrates efficient detection of hallucinated citation candidates in real-world scenarios, further supporting its practical utility.
Moreover, we evaluate efficiency by measuring the time required to verify citations in the first 100 long papers from ACL 2025. Experiments were conducted on three representative environments: MacBook Pro with Apple M3 chip, 24GB RAM, macOS 15.7.3, MacBook Air with Apple M1 chip, 8GB RAM, macOS 15.7.1 (standard student laptop provided by our institution), and AMD Ryzen 7 PRO 6850U (2.70 GHz), 32GB RAM, Windows 11 Pro with WSL (Ubuntu 22.04.4 LTS).
Table~\ref{tab:speed} shows that verification can be performed in at most 35 seconds per paper, and about 12 minutes for all 100 long papers. The primary bottleneck is the Citation Extractor component, indicating that further optimization of this stage would improve overall efficiency. Even on a MacBook Air, verification can be completed in approximately 15 seconds per paper, demonstrating the practical efficiency of the toolkit on commonly available hardware.

\section{Conclusion}

We introduced \textsc{HalluCiteChecker}, a toolkit for detecting hallucinated citations in scientific papers. We formalized citation verification as a structured task and presented a modular framework that decomposes the process into three subtasks, providing a clear foundation for future research. Moreover, as shown in Section~\ref{sec:evaluation}, the toolkit enables fine-grained measurement of processing time for each component, facilitating systematic analysis and optimization in future development.
Beyond its technical contributions, our goal is to support a more transparent and constructive scientific review process. By enabling systematic and reproducible citation verification, the toolkit can help reduce unintended desk rejections and unnecessary reviewer workload, thereby promoting responsible research practices among authors, reviewers, and organizers.

\section*{Ethics and Broader Impact Statements}

\paragraph{License.}
All tools, models, and datasets used in this work are distributed under permissive licenses, specifically the MIT License or the Apache 2.0 License. Accordingly, we release \textsc{HalluCiteChecker} under the Apache 2.0 License to ensure broad accessibility, reproducibility, and responsible reuse. The papers used for evaluation and examples were obtained from the ACL Anthology, which is distributed under the Creative Commons Attribution 4.0 International License (CC BY 4.0). Therefore, their use in this study complies with applicable licensing terms and poses no legal or ethical concerns.

\paragraph{Impact on reproducibility and archival transparency.}
This work primarily evaluates hallucinated citation detection using papers from ACL venues, but the toolkit is designed to generalize to other conferences such as ICLR and NeurIPS. However, unlike ACL Anthology, which maintains versioned archives of published papers, some venues overwrite manuscripts with updated versions without preserving historical revisions\footnote{\url{https://gptzero.me/news/iclr-2026/}, \url{https://gptzero.me/news/neurips/}}. This lack of versioned history makes independent verification of previously reported hallucinated citations difficult. We therefore emphasize the importance of transparent and versioned archival practices to support reproducibility, accountability, and long-term verification in scientific communication.

\paragraph{Impact of limited benchmarks and future research.}
Although concurrent work has begun to explore hallucinated citation detection, many existing tools and analyses are not easily reproducible, are environment-specific, or are not designed for general-purpose use across diverse publication formats. In contrast, \textsc{HalluCiteChecker} is designed with usability, reproducibility, and extensibility in mind, enabling broader adoption and systematic evaluation. Moreover, comprehensive benchmarks for hallucinated citation detection remain limited. We hope that this work provides a foundation for future dataset construction, benchmark development, and methodological improvements. By releasing both the task formulation and implementation, we aim to support further research in citation verification and scientific reliability. %

\paragraph{Responsible use and community impact.}
The use of AI-assisted writing tools is not inherently problematic; however, authors remain responsible for verifying the accuracy of all cited references. Hallucinated citations may arise when automatically generated suggestions are incorporated without sufficient verification. Our goal is not to discourage the use of AI-assisted research tools, but to provide a practical mechanism that enables authors, reviewers, and organizers to systematically verify citations and reduce unintended errors. This tool is intended to support a responsible research ecosystem by improving transparency, reducing unnecessary reviewer burden, and helping prevent avoidable desk rejections caused by citation errors.
If tools such as \textsc{HalluCiteChecker} become widely adopted within the community, similar to existing tools such as \textsc{ACL pubcheck}, we encourage community-driven maintenance and governance to ensure transparency, sustainability, and alignment with community needs. We hope this work contributes to establishing fair, transparent, and responsible scientific practices.

\bibliography{anthology-1, anthology-2, custom}

\end{document}